\documentclass[11pt]{article}

\usepackage[preprint]{acl}

\usepackage{times}
\usepackage{latexsym}
\usepackage{amsmath}
\usepackage{array}
\usepackage{booktabs}
\usepackage{float}
\usepackage{afterpage}
\usepackage{authblk}

\usepackage[T1]{fontenc}

\usepackage[utf8]{inputenc}

\usepackage{microtype}

\usepackage{inconsolata}

\usepackage{graphicx}

\title{Towards Human Cognition: Visual Context Guides Syntactic Priming in Fusion-Encoded Models}

\author[1]{Bushi Xiao}
\author[1]{Michael Bennie}
\author[1]{Jayetri Bardhan}
\author[1]{Daisy Zhe Wang}
\affil[1]{
  University of Florida\\
  \texttt{\{xiaobushi, michaelbennie, jayetri.bardhan\}@ufl.edu}, \texttt{daisyw@cise.ufl.edu}
}

\begin{document}
\maketitle
\begin{abstract}
Structural priming is a cognitive phenomenon where exposure to a particular syntactic structure increases the likelihood of producing the same structure in subsequent utterances. While humans consistently demonstrate structural priming effects across various linguistic contexts, it remains unclear whether multimodal large language models (MLLMs) exhibit similar syntactic preservation behaviors. We introduce PRISMATIC, the first multimodal structural priming dataset, which advances computational linguistics by providing a standardized benchmark for investigating syntax-vision interactions. We propose the Syntactic Preservation Index (SPI), a novel reference-free evaluation metric designed specifically to assess structural priming effects in sentence level. Using this metric, we constructed and tested models with two different multimodal encoding architectures to investigate their structural preservation capabilities. Our experimental results demonstrate that models with both encoding methods show comparable syntactic priming effects. However, only fusion-encoded models exhibit robust positive correlations between priming effects and visual similarity, suggesting a cognitive process more aligned with human psycholinguistic patterns. This work provides new insights into evaluating and understanding how syntactic information is processed in multimodal language models.
\end{abstract}

\section{Introduction}

Structural priming, first systematically studied by \citet{bock1986syntactic}, refers to the cognitive phenomenon where a syntactic structure increases the likelihood of producing the same structure in subsequent utterances. This phenomenon has profound implications for understanding human language processing mechanisms, as it reveals how syntactic representations are activated and reused during communication \citep{pickering2008structural}. Research has demonstrated that structural priming is pervasive across various linguistic contexts and provides crucial insights into the cognitive architecture underlying language production and comprehension. 

Recent advances in Large Language Models (LLMs) have sparked significant research interest across multiple domains \citep{app14052074}. Given the remarkable performance of these models in various linguistic tasks, researchers have begun investigating whether LLMs exhibit human-like structural priming effects \citep{sinclair-etal-2022-structural}. As \citet{zhao2023survey} noted, while initial studies focused on text-only scenarios, the integration of visual information processing has become a new research direction. This has catalyzed the emergence of Multimodal Large Language Models (MLLMs), which combine visual and textual information processing through various architectural approaches \citep{yin2023survey}. Early vision-language models like CLIP \citep{Radford2021LearningTV} and OFA \citep{pmlr-v162-wang22al} demonstrated different strategies for multimodal integration, laying the groundwork for more sophisticated MLLMs that combine billion-parameter language models with visual encoders. However, while research has examined structural understanding in MLLMs through contrastive evaluation and probing tasks \citep{nikolaus2022vision, DBLP:journals/corr/abs-2102-11115}, the investigation of structural priming in multimodal contexts remains largely unexplored. The prevalence of text-image applications in real-world scenarios demands a deeper understanding of how multimodal language models process and preserve syntactic structures when both visual and textual information influence structural choices during generation.

Given the remarkable performance of these models in various linguistic tasks, researchers have begun investigating whether LLMs exhibit human-like structural priming effects \citep{sinclair-etal-2022-structural}. As current large language models increasingly integrate visual information processing capabilities \citep{yin2023survey}, the study of structural priming in multimodal contexts has become particularly crucial. The prevalence of text-image applications in real-world scenarios demands a deeper understanding of how multimodal language models process and preserve syntactic structures when both visual and textual information are present.

\begin{figure}[h]
    \centering  
    \includegraphics[width=1.0\linewidth]{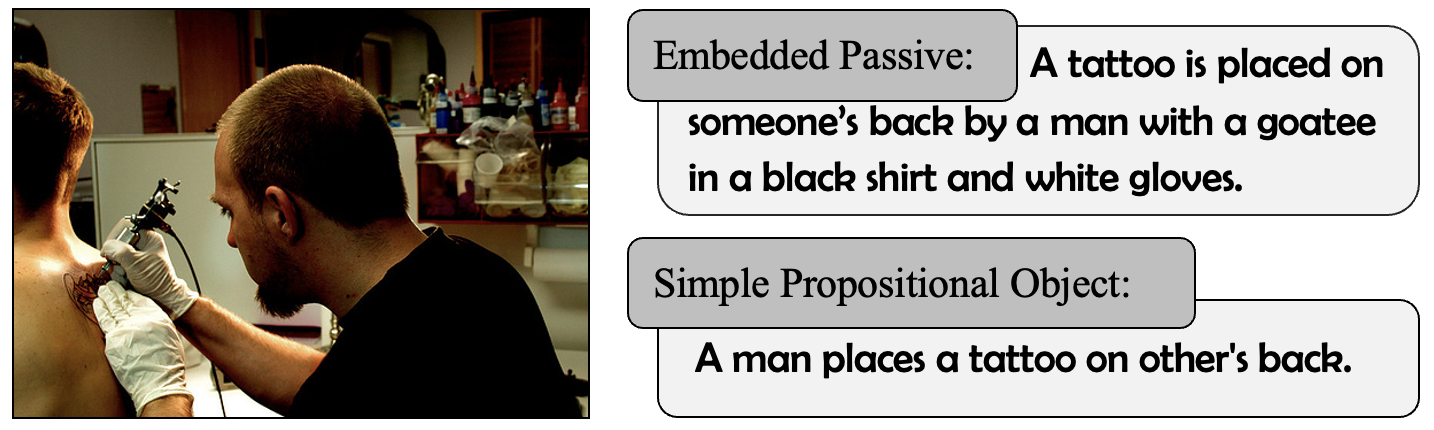}
    \caption{An example from the PRISMATIC dataset. Each image in the PRISMATIC dataset has multiple descriptions with the same semantics but different syntax.
    }
    \label{fig:dataset}
\end{figure}

However, multimodal structural priming remains largely unexplored, with no systematic studies or datasets addressing this phenomenon. To address this gap, we present PRISMATIC (PRIming through Syntactic Manipulation And Text-Image Coupling), a novel syntactic priming dataset derived from Flickr30k \citep{young-etal-2014-image}. The dataset comprises 4208 samples paired with aligned images. Figure \ref{fig:dataset} shows an example from our PRISMATIC dataset. An image is paired with multiple descriptions that share the same meaning but use different syntactic structures. This tuple presents a tattooing scene with two alternative descriptions as examples: one using embedded passive voice and another using simple propositional object construction. The dataset is constructed using a template-based methodology and validated by professional linguists to ensure quality. This resource is specifically designed to evaluate structural preservation capabilities in multimodal language models through various syntactic representations.

Alongside our dataset, we introduce the Syntactic Preservation Index (SPI), a novel reference-free evaluation metric specifically designed to assess structural priming effects without requiring predefined target sentences. Unlike previous evaluation methods that rely on probability prediction tasks with fixed targets, SPI enables more flexible and comprehensive evaluation of models' true generative capabilities in structural preservation.

We conducted controlled experiments comparing multiple architectural approaches, including dual-encoder models that process visual and semantic information separately, fusion-encoder models based on OFA-large \citep{pmlr-v162-wang22al}, and open-source multimodal language models such as LLaVA-v1.5-7B \citep{liu2023visualinstructiontuning} and BLIP-2-OPT-2.7b \citep{li2023blip2bootstrappinglanguageimagepretraining} \footnote{Code: 

\url{https://github.com/michaelbennieUFL/2025MLLM}
Data:

\url{https://github.com/kitayamachingtak/PRISMATIC}}.

Our findings reveal that multimodal language models exhibit human-like structural priming effects when priming elements are introduced, despite their different encoding mechanisms. This is contrary to the common inference that the dual encoding model has a weak effect on syntactic structure. Further correlation analysis between priming information similarity and SPI scores yielded a significant finding: fusion-encoded models demonstrate strong correlations between syntactic priming effects and visual content similarity. This aligns with human cognitive patterns, suggesting that fusion encoding may better approximate human information processing mechanisms, offering potential scientific insights into cognition mechanisms.
\\[5pt]
Our primary contributions are:
\begin{enumerate}
\item We created PRISMATIC, the first multimodal structural priming dataset with aligned image-sentence pairs, establishing a standardized benchmark that advances computational linguistics research in vision-language syntactic interactions.

\item We proposed Syntactic Preservation Index (SPI), a novel tree kernel-based evaluation metric that quantifies structural priming effects without requiring reference answers. SPI provides the first sentence-level measurement standard for both multimodal syntactic preservation with superior interpretability.

\item We conducted controlled experiments on dual and fusion encoding methods in MLLMs, revealing that fusion-encoded models exhibit more human-like cognitive patterns with stronger visual-syntactic correlations.
\end{enumerate}

\section{Related Works}
Since the emergence of Large Language Models (LLM) and MLLMs, most research has been focused on improving training efficiency, fine-tuning methods, few-shot learning, and thought chain capabilities \citep{app14052074, yin2023survey, liang2024survey}. However, some studies have specifically investigated the language structure in computational models \citep{huang2023structureclipscenegraphknowledge} . 

\subsection{Multimodal Large Language Models}
Although large language models are becoming more powerful with the development of self-attention transformer \citep{zhao2023survey}, they are still only able to recognize text information. To overcome this limitation, researchers have been exploring ways to combine visual and textual information processing.

Early efforts in vision-language models demonstrated various approaches to multimodal integration. CLIP \citep{Radford2021LearningTV} projects image and text as vectors into a unified representation space, while OFA \citep{pmlr-v162-wang22al} employs a unified transformer for joint encoding. Other models like ViLT \citep{kim2021viltvisionandlanguagetransformerconvolution} and LXMERT \citep{DBLP:journals/corr/abs-1908-07490} explore different architectural choices.

The emergence of MLLMs marks a significant advancement in this field. As summarized by \citet{yin2023survey}, MLLMs typically consist of three key components: a pre-trained encoder, a connector, and an LLM. Unlike traditional vision-language models, MLLMs are distinguished by their integration of billion-parameter language models and their ability to leverage multimodal instruction adaptation techniques. 

\subsection{Language Structures}
Language structural analyses in MLLMs can be broadly categorized into two main approaches: \citet{nikolaus2022vision} used contrastive evaluation with image-sentence pairs to test grammatical understanding, while \citet{DBLP:journals/corr/abs-2102-11115} employed probing tasks to show that structural information can be preserved during multimodal pre-training, though this depends on model design choices.

To enhance structural understanding, \citet{huang2023structureclipscenegraphknowledge} proposed Structure-CLIP, which incorporates explicit scene graph modeling to better preserve grammatical relationships. The importance of architectural choices in structural preservation was further confirmed by \citet{mckinzie2024mm1methodsanalysis}, who conducted comprehensive ablation studies showing the significant impact of image encoders on MLLM performance.

However, existing research primarily focuses on evaluating responses against predefined answers. These approaches assess whether models can correctly describe images or verify factual statements, but do not examine how visual information influences their structural choices in generation tasks. Such evaluation methods fail to capture the dynamic nature of language production, where multiple syntactic structures can be equally valid. These limitations suggest the need for more nuanced evaluation methods that consider both contextual processing and preference selection in structural understanding.

\subsection{Structural Priming}
 In human language processing, structural priming effects are well-attested in both comprehension and production \citep{tooley2014parity}. In particular, experiments have shown that ungrammatical and semantically incongruent sentences (e.g. "the waitress brings the book to the monk") elicit similar priming effects as well-formed sentences \citep{ivanova2017you}. This suggests that structural persistence effects are robust, even in the absence of semantic and lexical cues, providing insights into both language processing and machine communication \citep{linzen2021syntactic}.

In the field of computational linguistics, several studies have explored structural priming in language models. \citet{prasad2019using} introduced an Adaptation Effect metric to quantify structural similarities and demonstrated that trained LSTM models capture abstract language features beyond the word level. \citet{frank2019neural} showed that RNNs can preserve structural priming effects in monolingual contexts. Advancing this line of research, \citet{sinclair-etal-2022-structural} developed a new indicator to measure priming effects and created PRIME-LM, a corpus for various syntactic structures. \citet{michaelov_structural_2023} provided evidence that multilingual LLMs possess abstract syntactic representations that similarly affect text generation across languages. \citet{zhang2024modelingbilingualsentenceprocessing} revealed transformers outperform RNNs in cross-language priming. Most recently, \citet{jumelet2024language} tested the factors that influence the priming effect in LLMs, which proves that context also has an important influence on the syntactic structure of LLMs. \citet{tooley2025putting}'s research on humans also found that when the priming sentence and the target sentence share similar content, the processing relationship between the two is stronger.



\begin{figure*}[t]
    \centering  
    \includegraphics[width=1.0\linewidth]{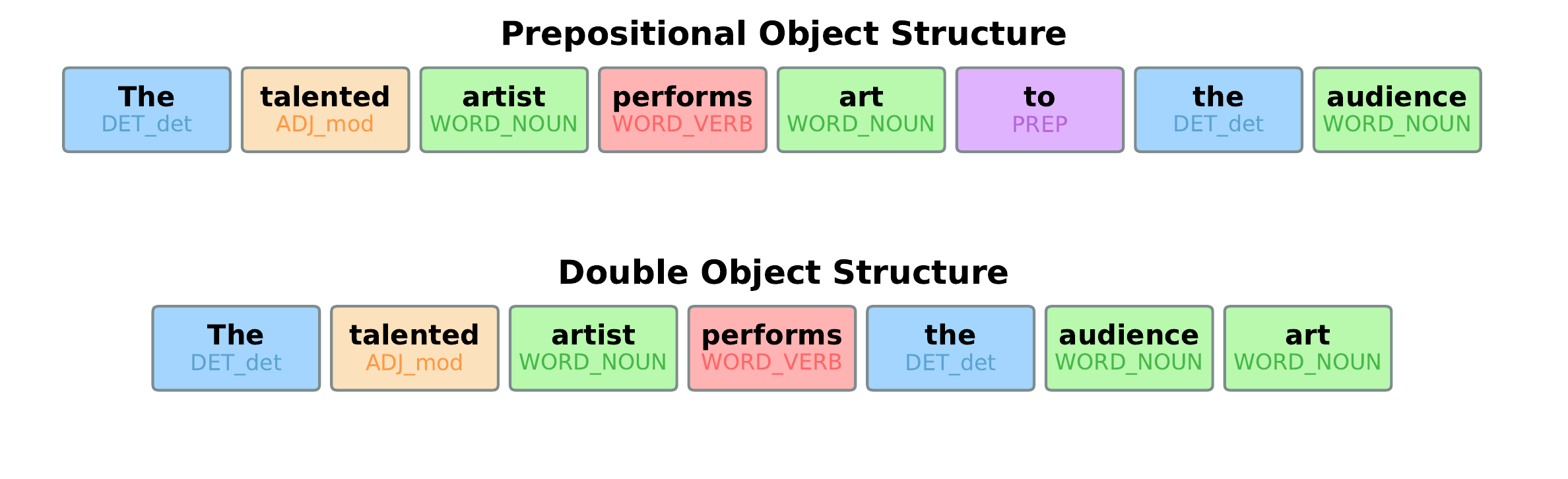}
    \caption{An example of the template for generating a priming pair: prepositional object sentence and double object sentence.}
    \label{fig:template}
\end{figure*}

\begin{table*}[t]
\caption{Syntactic Structure Types and Examples}
\small
\begin{tabular}{>{\raggedright\arraybackslash}p{2cm}>{\raggedright\arraybackslash}p{6cm}>{\raggedright\arraybackslash}p{7cm}}
\hline
Type & Structure & Example Sentence \\
\hline
Simple Active & \{subject\} \{verb\} \{obj\} & A boy carries a ball. \\
Simple Passive & \{subject\} \{auxiliary\_verb\} \{verb (past\_participle)\} \{by + agent\} & A ball is carried by a boy. \\
\hline
PO Passive & \{direct\_obj\} \{auxiliary\} \{past\_participle\} \{preposition\} by \{subject\} & The colors were painted on paper by a girl with the brush. \\
PO Active & \{subject\} \{active\_verb\} \{direct\_obj\} \{preposition\} \{prepositional\_object\} & A girl painted the colors on paper with the brush. \\
\hline
Embedded Passive & \{subject\} \{auxiliary\_verb\} \{verb (past participle)\} \{by + agent\} \{subordinate\_clause\} & The sidewalk that was washed by the women is green and purple. \\
Embedded Active & \{subject\} \{verb\} \{object\} \{subordinate\_clause\} & A woman washed the green and purple sidewalk. \\
\hline
Mediopassive & \{subject\} \{verb\} \{adverbial\_clause\} & The music plays loudly as the singer performs in front of the audience. \\
Mediopassive-like Active & \{subject\} \{verb\} \{adverbial\_modifier\} \{additional\_clause\} & The audience listens intently as the band plays their music. \\
\hline
Simple PO & \{subject\} \{verb\} \{direct\_object\} \{prep\} \{indirect\_object\} & A man tells stories to people. \\
Simple DO & \{subject\} \{verb\} \{indirect\_object\} \{direct\_object\} & A man tells people stories. \\
\hline
Complex PO & \{subject\_phrase\} \{verb\_phrase\} \{object\_phrase\}\{prep\_phrase\_text\} & A woman wearing black glasses share sweets with a toddler girl wearing a princess hat. \\
Complex DO & \{subject\_phrase\} \{verb\_phrase\} \{indirect\_object\_phrase\} \{direct\_object\_phrase\} & A woman wearing black glasses share a toddler girl wearing a princess hat sweets. \\
\hline
PO Clause & \{subject\} \{verb\}\{indirect\_object\_clause\} \{prep\} \{direct\_object\_clause\} & The teacher that carrys books give assignments to the student that studys in the library. \\
DO Clause & \{subject\} \{verb\} \{direct\_object\_clause\} \{indirect\_object\_clause\} & The teacher that carries books give the student that studys in the library assignments. \\
\hline
S-Genitive & \{possessor\} \{possessive 's\} \{possessed object\} & Reflections from the firefighters' uniforms. \\
Of-Genitive & \{possessed object\} \{of\} \{possessor\} & Reflections from the uniforms of the firefighters. \\
\hline
\end{tabular}
\label{tab:syntax-templates}

\end{table*}

\section{PRISMATIC Dataset}

PRISMATIC (PRIming through Syntactic Manipulation And Text-Image Coupling) comprises 4,208 manually curated and validated sentences spanning 16 syntactic priming conditions, aligned with 1,710 images. Each image is annotated with multiple descriptive sentences, with each sentence instantiating a specific syntactic structure. Notably, the distribution of syntactic conditions varies across images, as not all images are paired with all 16 structural variants. This dataset serves as a benchmark for evaluating visual-language models' capacity to integrate visual perception with syntactic processing.

\subsection{Dataset Construction}

The PRISMATIC dataset is built based on images and captions from Flickr30k \citep{young-etal-2014-image}, which contains 31,000 images with 5 caption sentences for each image.

\textbf{Reconstruct Syntax Trees:} The syntactic structure of each caption was converted to a syntax tree using the Natural Language Toolkit (NLTK) \citep{bird-loper-2004-nltk}. This identifies grammatical dependencies between words. Subsequently, each word of each syntax tree was assigned a tag that describes its syntactic role (e.g., subject, object, predicate).\footnote{Refer to Appendix \ref{appendix:syntax_tree} for an illustrative example of the syntax tree.}

\textbf{Fit into Templates:} This is the core step of dataset construction. For each original image description, we extract key semantic components and their syntactic roles from the parsed syntax tree. These labeled components are then systematically inserted into predefined syntactic templates to generate both positive and negative prime sentences.

As illustrated in Figure \ref{fig:template}, each template defines specific slots for different grammatical components (DET, ADJ, WORD\_NOUN, WORD\_VERB, PREP, etc.). For example, from sentence "The talented artist performs art to the audience.", each word is mapped to its corresponding template slot to create the prepositional object sentence construction. And the double object sentence became "The talented artist performs the audience art.". This template-based approach enables systematic generation of sentences with different syntactic structures while preserving semantic content.

Table \ref{tab:syntax-templates} presents the 16 syntactic types in our dataset with the corresponding tag combinations and examples. These structures fall into three major types: voice alternations (Active-Passive), dative alternations (PO-DO), and possessive alternations (S-Genitive/Of-Genitive), with multiple subcategories within each type.

\textbf{Sentence Filtering:} GPT-2 \citep{radford2019language} is used to calculate the perplexity \citep{jelinek1977perplexity}, which removes illogical sentences\footnote{Perplexity Threshold=300}.

\textbf{Grammatical Correction:} A fine-tuned version of the flan-t5-large model \citep{raheja2023coedittexteditingtaskspecific} was used to correct grammatical errors in the generated outputs.

\textbf{Human Annotation:} A human annotator who is a native English speaker made further corrections. See Section 3.3 for details.

\subsection{Automatic Evaluation}
To evaluate the quality of our dataset, we conducted a comprehensive automatic assessment using multiple metrics. Table \ref{tab:data-stats} presents the statistical analysis of our dataset.

\begin{table}[htbp]
\centering
\caption{Analysis of the Dataset}
\begin{tabular}{lr}
\hline
Feature & Value \\
\hline
Total Sentences & 4,208 \\
Total Words & 41,367 \\
Grammatical Error Rate & 1.06\% \\
\hline
Total Tokens & 48,350 \\
Word Types & 6293 \\
TTR & 0.1302 \\
Avg Tokens per Sentence & 11.49 \\
Token Range per Sentence & (4, 45) \\
\hline
Avg Perplexity & 86.37 \\
Perplexity Range & (7.02, 289.63) \\
\hline
\end{tabular}
\label{tab:data-stats}
\end{table}

We calculated the Type-Token Ratio (TTR) \citep{richards1987type} to measure vocabulary richness, which represents the ratio of unique words (word types) to total words (tokens). A TTR of 0.1302 indicates reasonable lexical diversity in our dataset.

Perplexity scores \citep{jelinek1977perplexity} were computed using GPT-2 to assess sentence naturalness and fluency. Lower perplexity values indicate more natural-sounding sentences. Our dataset achieves an average perplexity of 86.37, with values ranging from 7.02 to 289.63, suggesting that most generated sentences maintain reasonable fluency.

We utilized the LanguageTool Python library 2.8.2\footnote{https://pypi.org/project/language-tool-python/} to automatically detect grammatical errors. The evaluation revealed a grammatical error rate of 1.06\%, indicating high grammatical quality in the generated sentences.

The dataset contains 4,208 sentences with 48,350 total tokens, averaging 11.49 tokens per sentence with lengths ranging from 4 to 45 tokens. 


\subsection{Human Evaluation}
Given the limitations of automatic evaluation for assessing multimodal dataset quality, professional linguists conducted comprehensive human evaluation and refinement of the initially generated 4,896 sentences. The evaluation was performed across three critical dimensions:

1. Assessing whether generated sentences accurately describe the visual content of corresponding images (Error rate: 13.97\%)

2. Structural Alignment: Verifying that sentences correctly implement the intended syntactic structures as specified by their labels (Error rate: 4.41\%)

3. Grammatical Accuracy: Identifying and correcting grammatical errors in generated sentences (Error Rate: 10.29\%)

Based on this comprehensive evaluation, 688 sentences with severe errors were removed from the dataset to ensure data quality. Sentences with minor issues were corrected by the linguistic experts while preserving the intended syntactic structures and semantic content. This human validation process ensures that our dataset maintains high standards for both linguistic accuracy and multimodal alignment.

\section{Syntactic Preservation Index}
\begin{figure}[h]
    \centering  
    \includegraphics[width=\linewidth]{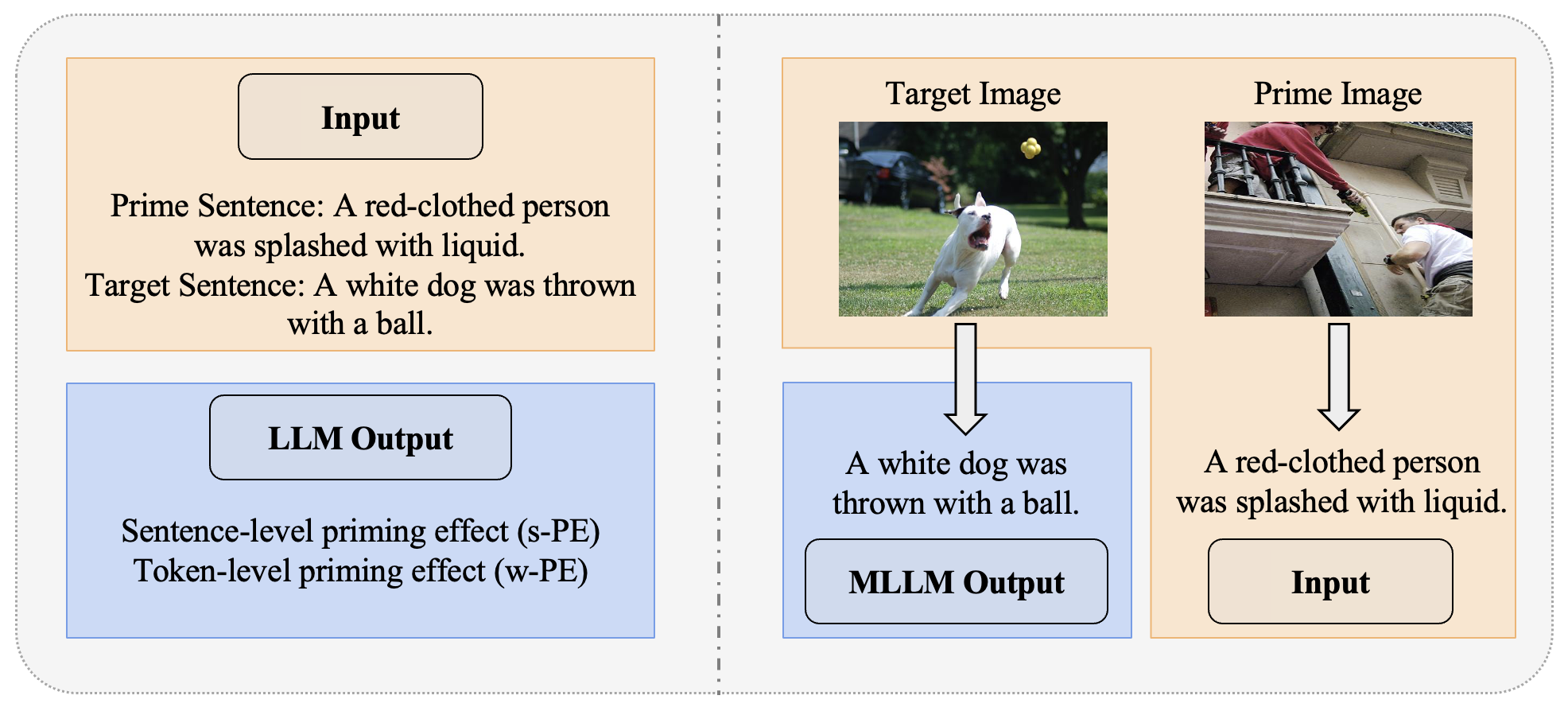}
    \caption{Comparison of evaluation methods for structural priming effect on LLMs. Left: Previous method proposed by \citet{prasad2019using}. Right: Our new MLLM model to produce target sentence directly.}
    \label{fig:comparison}
\end{figure}

Traditional evaluation methods for syntactic priming typically involve simultaneous input of both prime and target sentences into language models. The left panel of Figure \ref{fig:comparison} illustrates previous studies that evaluate the priming effect based on token probability \cite{prasad2019using, sinclair-etal-2022-structural, michaelov_structural_2023, jumelet2024language}. In that scenario, both the prime sentence and the Target Sentence are predetermined inputs, while LLM outputs the surprisal (token probability) to indicate the priming level. Although this framework is concrete and simple, it does not examine the model's ability to generate complete sentences. 

Our fairer evaluation method is to input the semantic information of the priming sentence and the visual information that aligns with the target sentence at the same time. As shown in the right panel of Figure \ref{fig:comparison}, the MLLM is required to predict the target sentence. Therefore, we propose a new metric based on the tree kernel algorithm \citep{moschitti2006making}. Any machine-predicted sentence will be directly compared with the prime sentence to get the SPI.

\subsection{Tree Kernel}
The tree kernel method \citep{moschitti2006making} calculates the structural similarity between two sentences by comparing their syntax trees \citep{DBLP:journals/corr/TaiSM15}. Given two trees $T_1$ and $T_2$, their Tree Kernel is defined as:
\begin{equation}
   K(T_1,T_2) = \sum_{n_1 \in N_1} \sum_{n_2 \in N_2} \Delta(n_1,n_2)
\end{equation}
where $N_1$ and $N_2$ are the sets of nodes in trees $T_1$ and $T_2$ respectively, and $\Delta(n_1,n_2)$ represents the number of common substructures between subtrees rooted at nodes $n_1$ and $n_2$.

The function $\Delta(n_1,n_2)$ is calculated as:
\begin{equation}
   \Delta(n_1,n_2) = 
   \begin{cases}
       0 & \text{if } prod(n_1) \neq prod(n_2) \\[2ex]
       \lambda & \text{if } pre-terminal(n_1) \\[2ex]
       \lambda \prod\limits_{j=1}^{nc(n_1)} & (1 + \Delta(ch(n_1,j), \\
       & ch(n_2,j))) \text{ otherwise}
   \end{cases}
\end{equation}
where:
\begin{itemize}
   \item $prod(n)$ is the production rule at node $n$
   \item $pre-terminal(n)$ determines if node $n$ is pre-terminal
   \item $nc(n)$ is the number of children of node $n$
   \item $ch(n,j)$ is the $j$-th child of node $n$
   \item $\lambda$ is a decay factor $(0 < \lambda \leq 1)$
\end{itemize}

To obtain a normalized similarity score, we use:
\begin{equation}
   K_{norm}(T_1,T_2) = \frac{K(T_1,T_2)}{\sqrt{K(T_1,T_1) \cdot K(T_2,T_2)}}
\end{equation}
Finally, the tree distance can be derived from the kernel function:
\begin{equation}
   d(T_1,T_2) = \sqrt{2 - 2K_{norm}(T_1,T_2)}
\end{equation}
This method effectively captures structural relationships in sentences and provides interpretable similarity measures without considering semantic difference.

\subsection{Syntactic Preservation Index}
Our proposed metric – SPI can be used to evaluate the intensity of structural priming. Let us consider a priming pair that describes the same picture:

1. Prepositional Object (PO): The talented artist performs street art for the audience.

2. Double Object (DO): The talented artist performs the audience street art.

For our experiment, we select the PO sentence as the positive prime sentence and the DO sentence as the negative prime sentence for comparison. Only the positive prime sentence is inputted along with a randomly selected target image from our dataset, then a predicted sentence is generated to describe the target image. 
\\[5pt]
\textbf{Notation and Definitions:}

- PP (Positive Prime): The syntax tree of the input sentence, which is the PO sentence

- NP (Negative Prime): The syntax tree of the corresponding sentence, which is the DO sentence here

- PS (Predicted Sentence): The syntax tree of the output sentence generated for the Target Image
\\[5pt]
\textbf{Algorithm Steps:}

The tree kernel between PP and PS:
\begin{equation}
D_p = K_{norm}(T_{pp}, T_{ps})
\end{equation}

The tree kernel between NP and PS:
\begin{equation}
D_n = K_{norm}(T_{np}, T_{ps})
\end{equation}
Here, $K(\cdot,\cdot)$ represents the tree kernel function that measures structural similarity between two syntax trees.

\begin{figure}[htbp]
    \centering  
    \includegraphics[width=\linewidth]{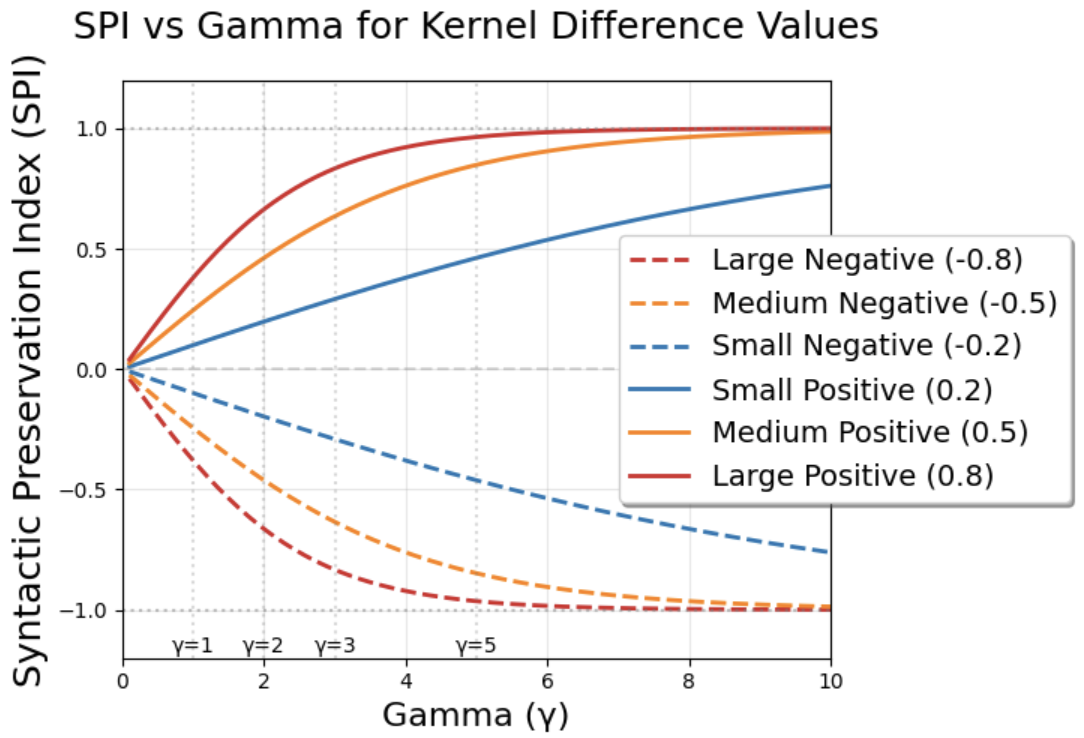}
    \caption{Relations of $\gamma$ value, SPI value and the kernel difference values.
    }
    \label{fig:gamma}
\end{figure}

We employed a normalized exponential amplification method to map the relative difference between these kernel values to [-1, 1]. When the predicted sentence structure is more similar to the positive priming, the value approaches 1; when it is more similar to the negative priming, the value approaches -1. This relationship is expressed as:
\begin{equation}
SPI= \frac{e^{\gamma(D_p-D_n)} - 1}{e^{\gamma(D_n-D_p)} + 1}, \quad 0.1 \leq \gamma \leq 10.0
\end{equation}
where $\gamma$ is a scaling factor that controls the sensitivity of the transformation. 
In Figure \ref{fig:gamma}, we have plotted SPI versus Gamma for different kernel values.  We examined different $\gamma$ values that result in different sensitivities. From the plot, we observe that that they eventually converge to 1 or -1.

\subsection{Metrics Comparison}
\begin{figure}[ht]
    \centering  
    \includegraphics[width=\linewidth]{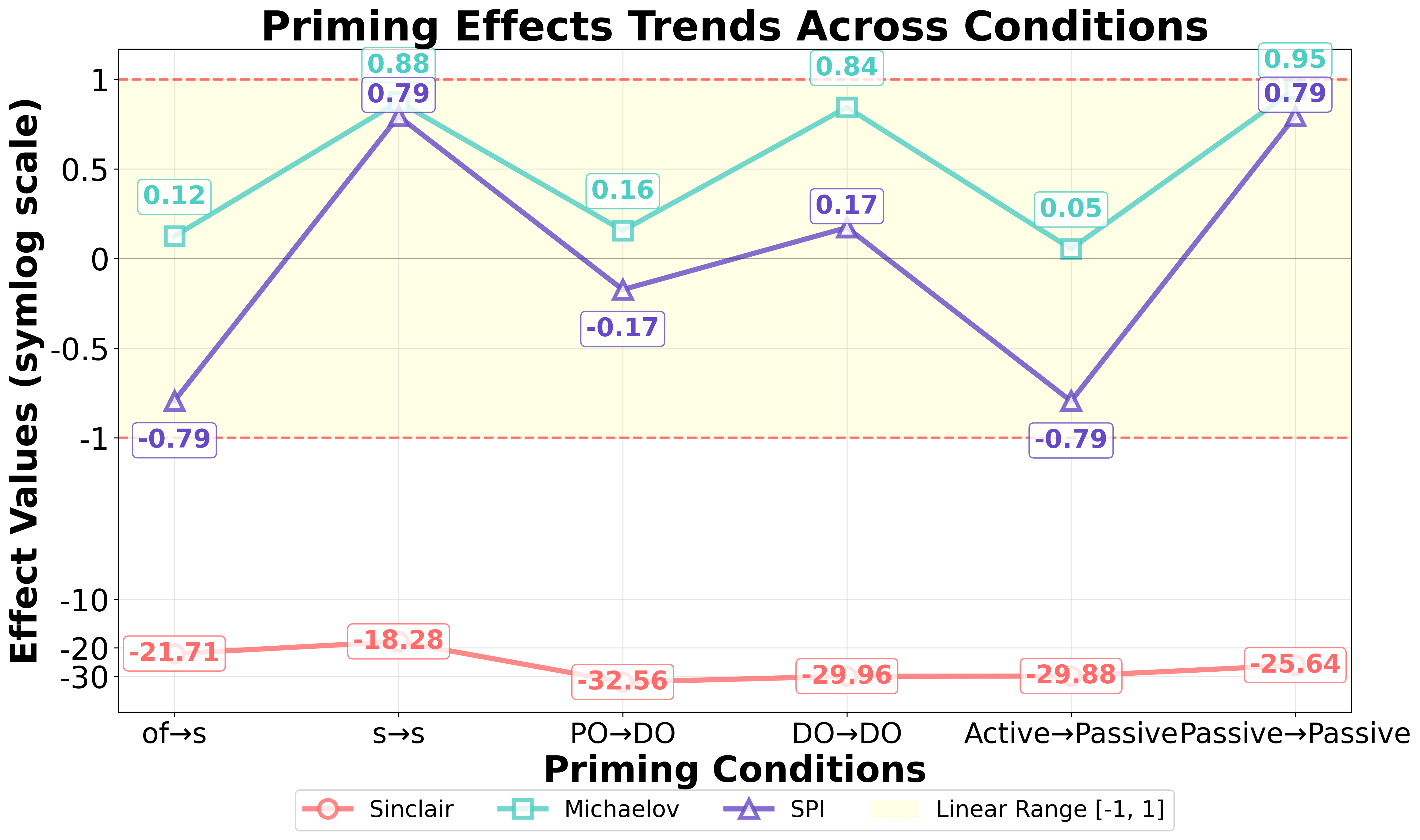}
    \caption{Experiment with SPI metric and other metrics.
    }
    \label{fig:metrics}
\end{figure}

To evaluate the effectiveness of our proposed SPI metric, we conducted comparative experiments using established datasets and baseline methods. We employed three syntactic priming datasets on three pairs of different priming types: \citet{Bernolet2013} for S-genitive and of-genitive, \citet{SCHOONBAERT2007153} for prepositional object and double object, and \citet{Kotzochampou2022} for active and passive. Since these datasets contain both multilingual data and English, we filtered them to retain only English priming pairs.

For baseline comparisons, we implemented two established structural priming metrics. First, we calculated surprisal values for tokens using GPT-2 \citep{radford2019language} to compute the Michaelov method \citep{michaelov_structural_2023} and the Sinclair method \cite{sinclair-etal-2022-structural}. For each target sentence, we also calculated SPI scores by comparing it directly against both positive prime and negative prime sentences.



Figure \ref{fig:metrics} presents the comparative performance of three structural priming metrics across different syntactic types. All three metrics demonstrate consistent trends, validating the feasibility of our proposed SPI metric. When prime and target sentences share the same structure, all metrics show high scores indicating strong priming effects; when they differ, scores are relatively low or negative.

The SPI metric offers several advantages over existing methods. It provides clear interpretability with positive values indicating priming facilitation and a zero threshold for effect detection. Notably, SPI exhibits perfect symmetry (e.g., of-s = -0.79, s-s = +0.79), establishing a unified absolute standard that enables evaluation of isolated priming pairs without requiring contrastive conditions.

Cross-construction analysis reveals that PO/DO alternations show weaker priming effects compared to genitive and voice alternations across all metrics. This pattern, most pronounced in SPI, reflects the syntactic complexity of argument structure alternations. 

More importantly, SPI is reference-free, allowing direct evaluation of any sentence generated by a large model. In contrast, all existing metrics require both the “answer” and the model input, computing token-level probabilities. In other words, SPI evaluates the generated output itself, while other metrics evaluate the probability of fixed input priming pairs, thus failing to capture the model’s true generative ability. Moreover, unlike text-only metrics, SPI is the only one can assess both purely textual and vision-language generations.

\section{Experiments}
We impose the following constraints on open-source MLLM selection. The model must exhibit: (i) support for continuous multimodal data processing, (ii) mechanisms for semantic relationship modeling, (iii) instruction-following capabilities through prompt understanding, and (iv) the ability to generate descriptions with temporal and contextual awareness beyond instantaneous visual observations.

\begin{figure}[t]
    \centering  
    \includegraphics[width=0.85\linewidth]{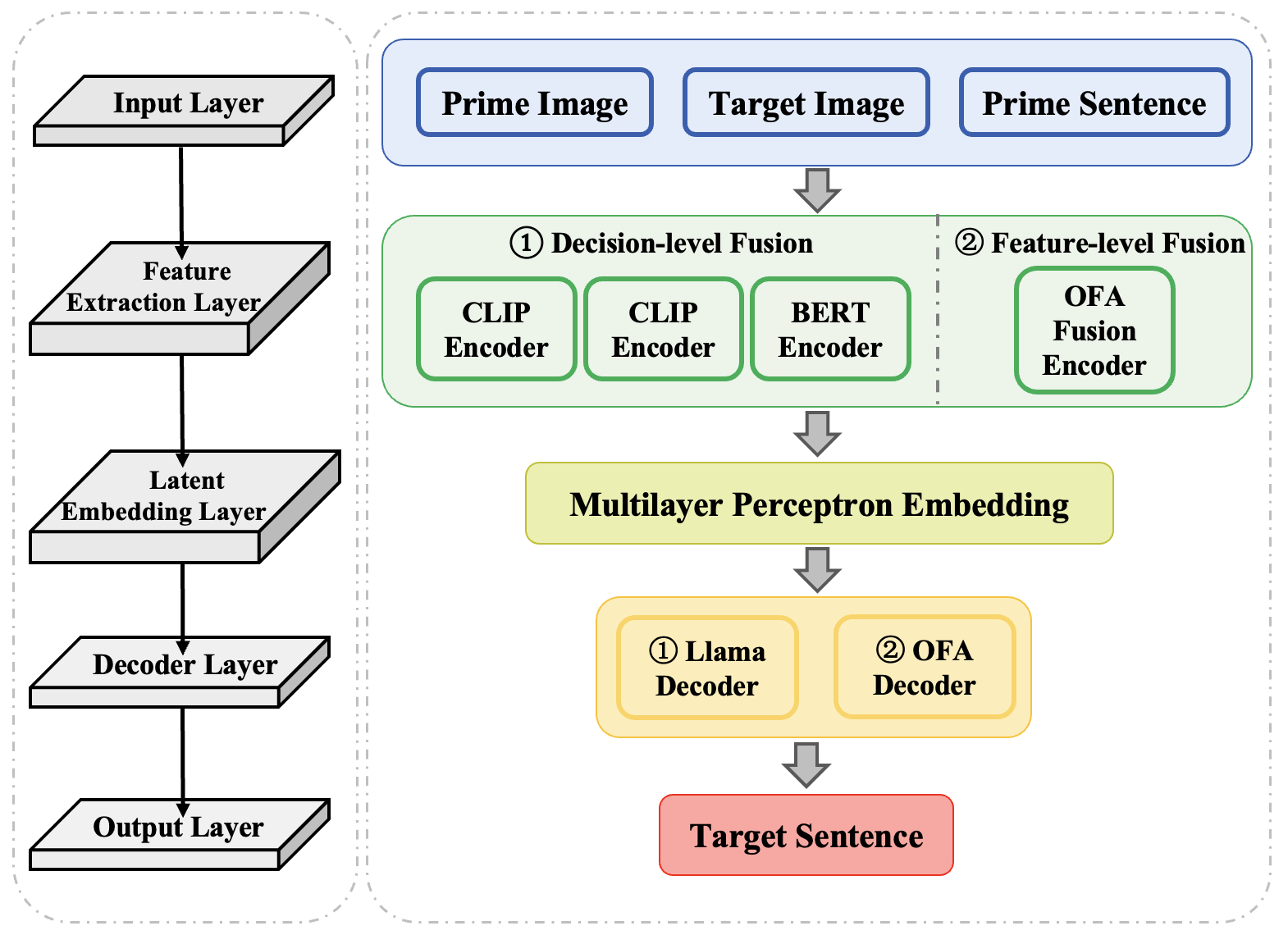}
    \caption{Structure comparison of Model 1 (left) using Dual Encoding and Model 2 (right) using Fusion Encoding.}
    \label{fig:structure}
\end{figure}

We selected two open-source MLLMs, LLaVA (fusion encoding) \citep{liu2023visualinstructiontuning} and BLIP-2 (dual encoding) \citep{li2023blip2bootstrappinglanguageimagepretraining} to examine their structural priming capabilities. Since they only support single-image input, the prime sentence and target image were inputted for a fair comparison. To enable more comprehensive analysis, we implemented our own architectures capable of processing a prime image, a prime sentence, and a target image simultaneously. For controlled experiments, we developed two MLLMs shown in Figure \ref{fig:structure}: a dual model integrating an MLP with a Llama decoder and a fusion model based on an OFA encoder.

\subsection{Model 1: Dual Encoding}
The dual encoding model utilizes BERT \citep{DBLP:journals/corr/abs-1810-04805} for text processing and CLIP \citep{Radford2021LearningTV} for image feature extraction. Multi-threading is employed for efficient data processing. The MLP module combines image and text embeddings, with a GELU for activation. They are then passed to a TinyLlama-1.1B decoder for natural language generation.

\subsection{Model 2: Fusion Encoding}
We trained the OFA encoder and used the same hyperparameters and dataset for fair comparison. Our model jointly encodes both images and fuses their representations with the prime sentence embedding. The encoder follows a transformer-based architecture, incorporating Gaussian Error Linear Units (GELU) activation for improved nonlinearity. ResNet-101 is used for visual feature extraction, ensuring robust image representations. These fused embeddings are then passed to the OFA decoder to generate the description of the target image.

\subsection{Training and Experimental Setup}

Both models were trained on COCO \citep{lin2014microsoft} using identical hyperparameters, including a batch size of 13, a tokenizer max length of 256, and same number of embedding dimensions. Model 1 was trained on a NVIDIA RTX 4060 GPU with an average training time of 57 minutes per epoch, while Model 2 was trained on a NVIDIA A100 GPU with an average training time of 50 minutes per epoch. The training process consisted of 9 epochs for both models.

We selected 1,006 annotated sentences across 16 syntactic types from PRISMATIC as the test set to avoid image duplication. Given a prime sentence and prime image, the program randomly selects a target image from the dataset, and the model generates a predicted sentence. Given the stochastic nature of target image selection, we employ our reference-free SPI metric rather than comparing against predetermined ground truth. Each sentence serves as a priming probe, and the process is repeated for 10 iterations, with a new target image randomly drawn in each iteration to assess the priming effect\footnote{Temperature set to 0.7 for all groups}.

We evaluated the following models:
\begin{enumerate}
\item LLaVA-1.5: A pre-trained fusion-encoded model.
\item BLIP-2: A pre-trained dual model.
\item Model 1: Our transparent dual encoding architecture.
\item Model 2: Our transparent fusion encoding architecture.
\end{enumerate}
Each experimental group was paired with a control group where no priming sentence or priming image was provided as input.

\section{Results}
\subsection{Image-Sentence Alignment}
We leveraged CLIP cosine similarity \citep{Radford2021LearningTV} to compute the semantic alignment between the images and their corresponding generated sentences. Sentence quality remains stable across different priming conditions. We observed that complex priming sentences slightly increase the similarity scores, suggesting that they encourage more detailed image descriptions.

\begin{table}[h]
    \centering
    \resizebox{1\linewidth}{!}{ 
    \begin{tabular}{lcccc}
        \toprule
        & LLaVA & BLIP-2 & Model 1 & Model 2 \\
        \midrule
        With Prime & 29.70 & 30.72 & 22.94 & 21.91 \\
        Without Prime & 32.82 & 30.08 & 22.37 & 22.76 \\
        \bottomrule
    \end{tabular}
    }
    \caption{Average CLIP cosine similarity scores for image-text matching across different models. Values represent percentage of matching scores.}
    \label{tab:clip-scores}
\end{table}

As depicted in table \ref{tab:clip-scores}, the CLIP similarity scores remain relatively consistent between primed and non-primed conditions. However, we observe different patterns across architectures. While dual models (BLIP-2 and Model 1) maintain stable performance, fusion-encoded models (LLaVA and Model 2) show decreased similarity scores when prime information is added. This suggests that fusion models are more sensitive to priming input, sometimes leading to descriptions that deviate from the image content or exhibit hallucination effects. Since LLaVA and BLIP-2 have been adjusted and optimized for diverse data sets, their overall score is higher.

\subsection{Syntactic Preservation Index}

We quantified performance using two metrics: SPI \footnote{$\gamma$=3} and structural preservation rate (the proportion of generated sentences that maintain the prime sentence's syntactic structure). Intra-model comparisons reveal significant improvements in SPI when priming sentences are provided as input compared to non-primed conditions. 

\begin{figure*}[t]
    \centering  
    \includegraphics[width=\linewidth]{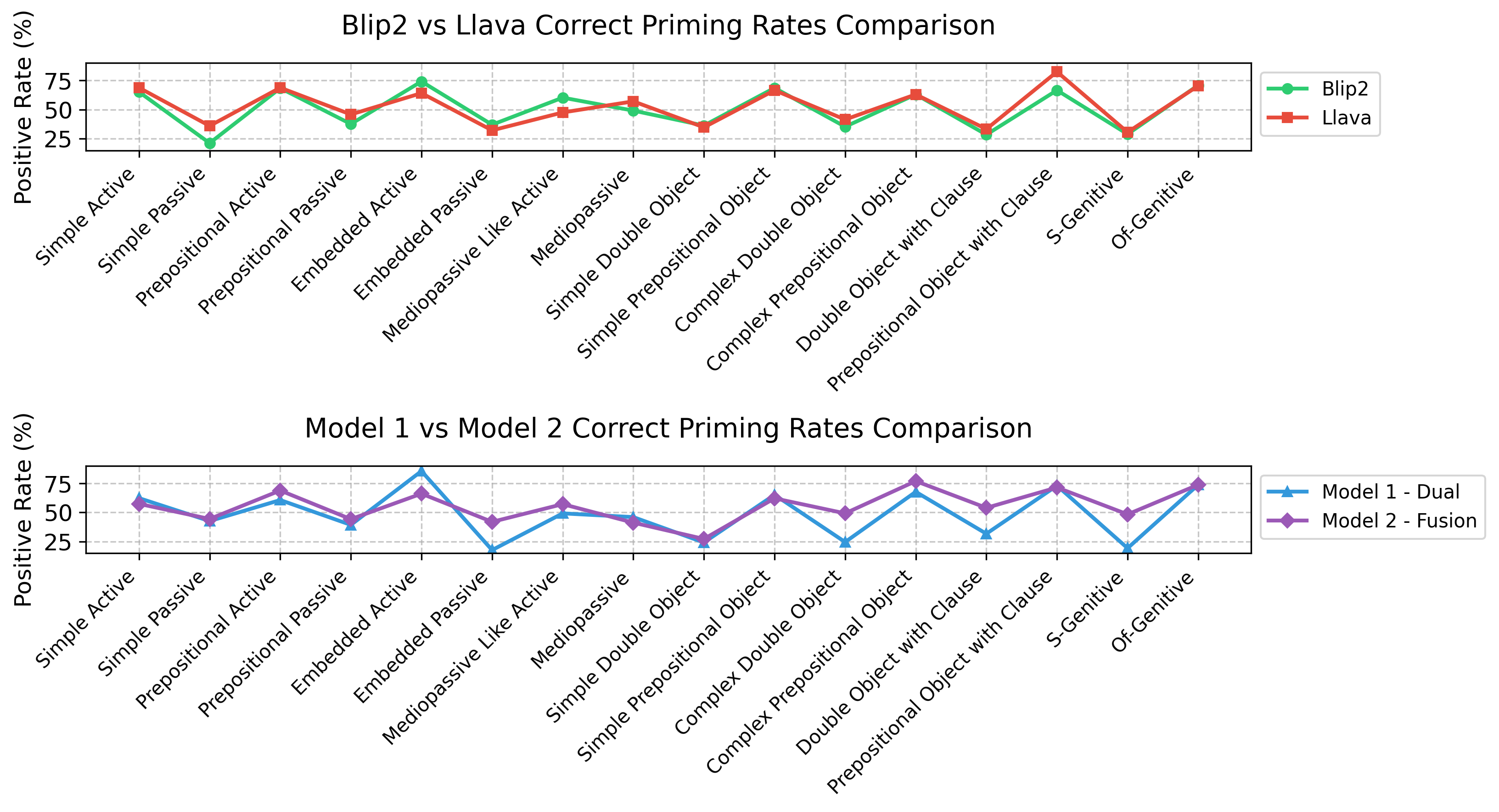}
    \caption{Comparison of the Correct (Positive) Rate of all models under different priming types.}
    \label{fig:all_models}
\end{figure*}

\begin{figure}[h]
    \centering  
    \includegraphics[width=\linewidth]{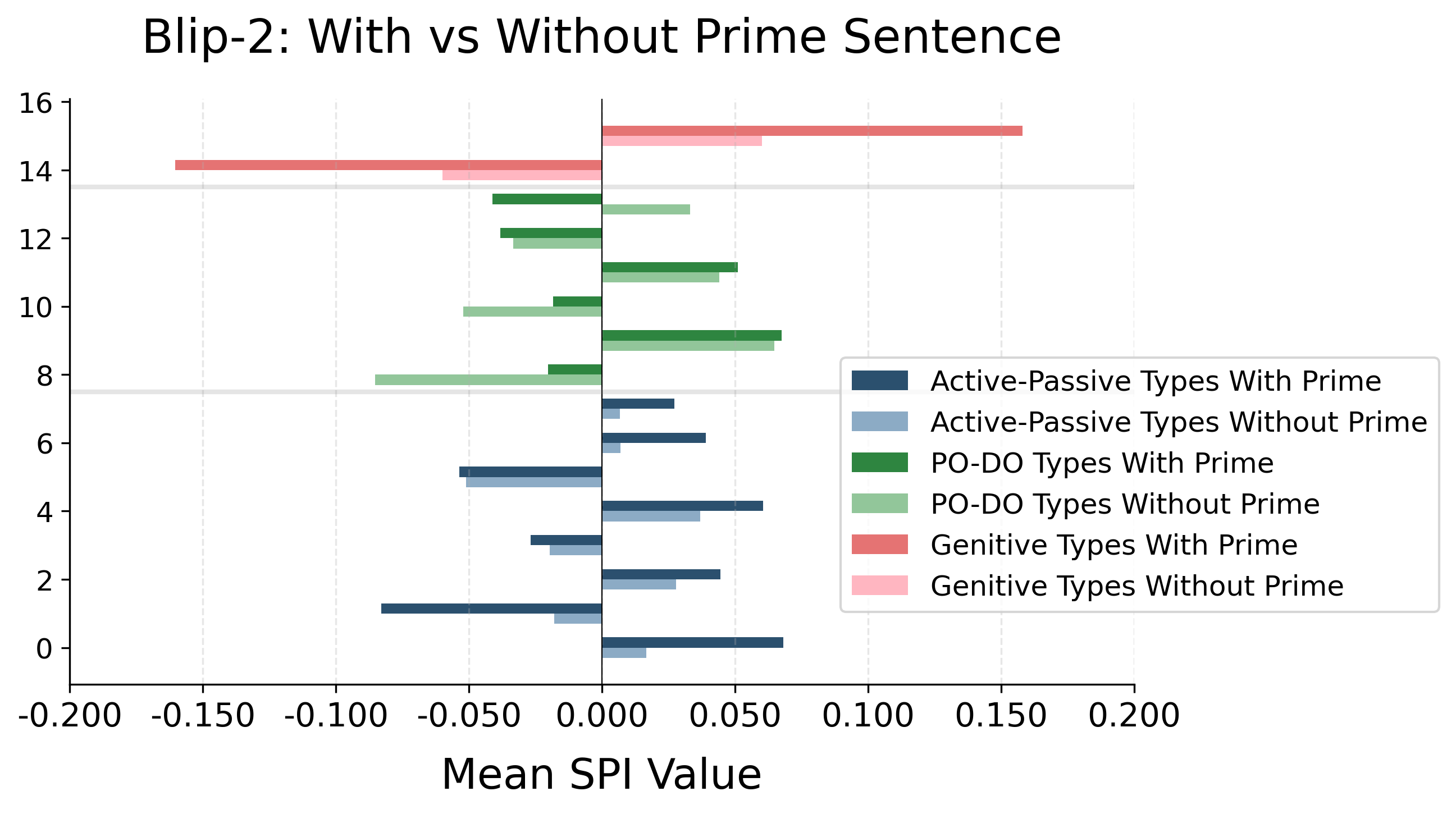}
    \caption{Comparative analysis of BLIP-2's SPIs under primed and non-primed conditions.}
    \label{fig:blip_prime}
\end{figure}

Take the figure \ref{fig:blip_prime} as an example, BLIP-2 demonstrates clear structural priming effects across all three syntactic categories when prime sentences are provided. Contrary to theoretical predictions suggesting dual-encoding architectures would exhibit minimal priming effects due to their independent visual and textual processing pathways, our results reveal systematic improvements in structural preservation. Specifically, Active-Passive constructions show enhanced structural alignment with prime sentences (positive SPI values ranging from approximately -0.05 to 0.05 with priming versus -0.08 to 0.05 without), while PO-DO (Prepositional Object-Direct Object) types exhibit the most pronounced priming benefits, with consistently higher SPI values when primed. Genitive constructions maintain stable positive preservation scores regardless of priming condition, but show slight improvements with prime sentence input. These findings suggest that even in dual-encoded architectures, cross-modal structural influences emerge during the generation phase, challenging assumptions about the rigid independence of processing streams in such models.

\begin{figure*}[t]
    \centering  
    \includegraphics[width=\linewidth]{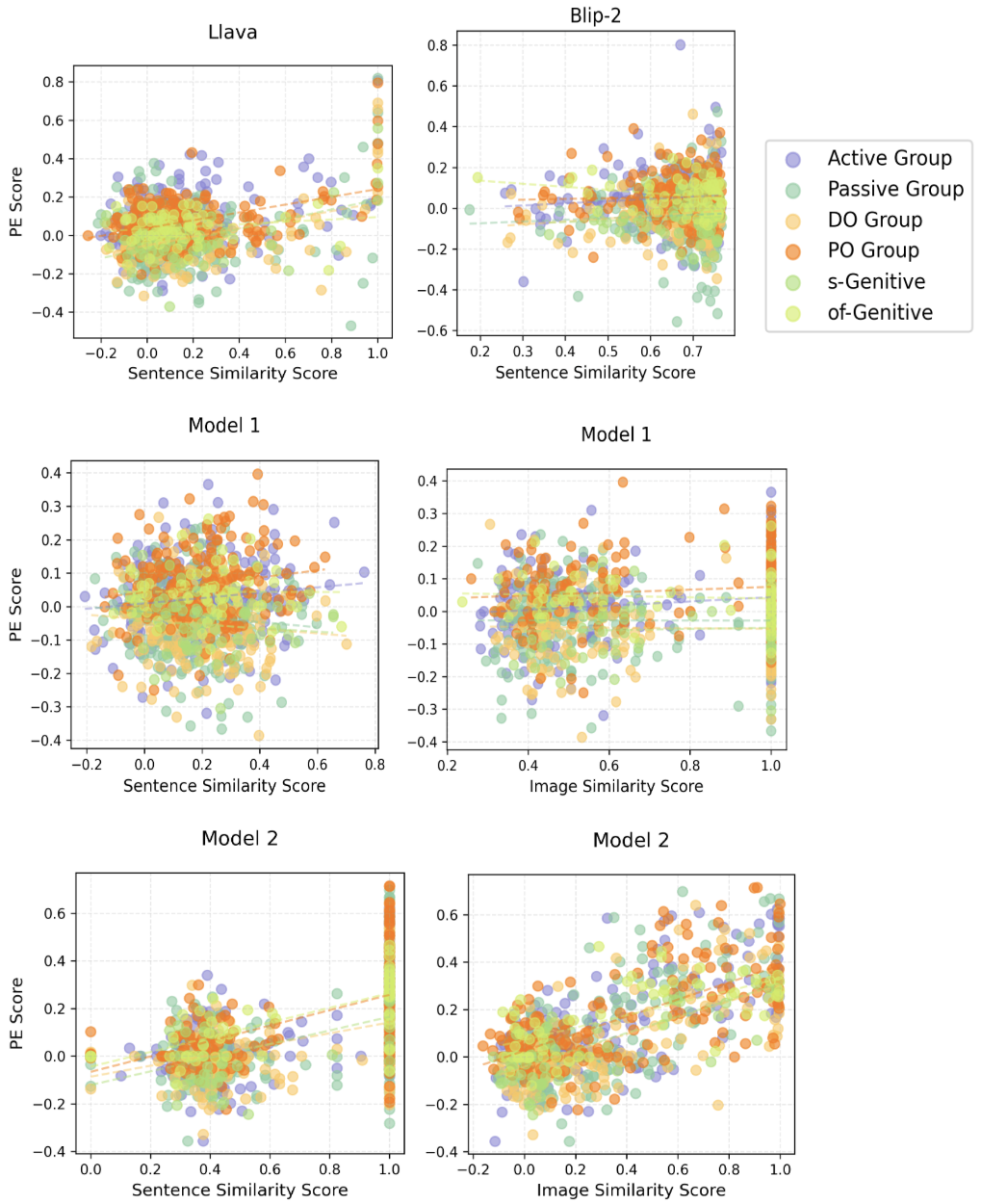}
    \caption{The correlation between sentence similarity or image similarity to the SPI.}
    \label{fig:correlations}
\end{figure*}

Figure \ref{fig:all_models} presents a comprehensive comparison of structural preservation rates across different syntactic constructions. The upper panel reveals that while BLIP-2 and LLaVA exhibit comparable performance across most syntactic types, notable differences emerge in specific constructions: BLIP-2 shows superior preservation in Simple Active and Prepositional Active structures (reaching approximately 65\% positive rates), whereas LLaVA demonstrates stronger performance in Complex Prepositional Object constructions and S-Genitive structures (achieving over 70\% positive rates). The lower panel comparing our custom architectures shows that Model 1 (dual encoding) and Model 2 (fusion encoding) maintain remarkably similar performance patterns across most syntactic categories, with positive rates consistently ranging between 25-75\%. However, Model 2 exhibits slightly more stable performance, particularly in embedded constructions and complex syntactic types, suggesting that fusion encoding may provide more robust structural preservation capabilities in challenging syntactic contexts. Importantly, all four models demonstrate positive structural priming effects across the majority of syntactic constructions, indicating that multimodal structural priming is a consistent phenomenon regardless of architectural approach.

\subsection{Correlation with Context Similarity}

We use OpenCLIP \citep{Radford2021LearningTV} to calculate the visual similarity between prime images and target images and use Sentence Transformers \citep{reimers-2019-sentence-bert} to calculate the semantic similarity between Prime sentences and Target. We then performed a correlation analysis between the two arrays using the Pearson correlation coefficient:

\begin{equation}
r = \frac{\sum_{i=1}^{n} (x_i - \bar{x})(y_i - \bar{y})}{\sqrt{\sum_{i=1}^{n} (x_i - \bar{x})^2} \sqrt{\sum_{i=1}^{n} (y_i - \bar{y})^2}}
\end{equation}

Our correlation analysis aligns with and extends previous findings in both computational and psycholinguistic research \citep{jumelet2024language, tooley2025putting}, revealing distinct patterns across different architectural approaches. As shown in Table \ref{tab:correlation}, fusion-encoded models demonstrate significantly stronger correlations with sentence semantic similarity compared to dual-encoded models. Specifically, LLaVA exhibits a moderate positive correlation (r = 0.3212, p = 1.58e-25), while BLIP-2 shows virtually no correlation (r = 0.0358, p = 0.2595). This architectural divide becomes even more pronounced in our custom models: Model 2 (fusion encoding) achieves a robust correlation of r = 0.5745 (p = 2.99e-89), whereas Model 1 (dual encoding) shows no significant relationship (r = 0.0295, p = 0.3500).

\begin{table}[h]
    \centering
    \resizebox{\linewidth}{!}{ 
    \begin{tabular}{lcc}
        \toprule
        & LLaVA & BLIP-2 \\
        \midrule
        Sentence Correlation coefficient & 0.3212 & 0.0358 \\
        P-value & 1.58e-25 & 0.2595 \\
        \midrule
        & Model 1 & Model 2 \\
        \midrule
        Sentence Correlation coefficient & 0.0295 & 0.5745 \\
        P-value & 0.3500 & 2.99e-89 \\
        \hline
        Image Correlation coefficient & 0.0729 & 0.7018 \\
        P-value & 0.0207 & 8.89e-150 \\ 
        \bottomrule
    \end{tabular}
    }
    \caption{Correlation coefficient across all models.}
    \label{tab:correlation}
\end{table}

The visual similarity analysis further reinforces this architectural distinction. Model 2 demonstrates an exceptionally strong correlation between image similarity and SPI (r = 0.7018, p = 8.89e-150), while Model 1 shows only a weak correlation (r = 0.0729, p = 0.0207). Remarkably, the correlation between visual similarity and structural priming (r = 0.7018) substantially exceeds that of sentence semantic similarity (r = 0.5745) in Model 2, suggesting that visual stimuli exert stronger influence on syntactic preservation than linguistic similarity alone. This finding challenges traditional assumptions about language processing hierarchies and indicates that visual context may serve as a more stable cognitive anchor for structural priming than semantic content, particularly in fusion-encoded architectures that enable deeper multimodal integration. 

Figure \ref{fig:correlations} presents 1,000 randomly sampled data points illustrating the relationship between similarity scores (sentence or image) and SPI scores across different models. For BLIP-2 and Model 1, the distributions exhibit high dispersion with no obvious correlation pattern. Also, their data are mostly clustered at the center of the graph. In contrast, LLaVA and Model 2 demonstrate clear positive trends: as similarity scores increase, SPI scores correspondingly rise, forming approximate linear relationships. This correlation is most pronounced in Model 2's image similarity condition, where the data points form a distinct upward trajectory.

\section{Discussion}

The superior correlation patterns in fusion models suggest that true multimodal structural priming requires unified representational spaces where visual and linguistic information can mutually influence syntactic decisions. This integration mechanism aligns with psycholinguistic evidence that visual context directly influences real-time language processing. \citet{ALTMANN2007502} showed humans make anticipatory eye movements based on tense—looking toward empty wine glasses for "has drunk" versus full beer glasses for "will drink"—demonstrating that language processing is fundamentally grounded in visual affordances. We hypothesize that fusion encoding creates shared semantic-syntactic mappings that allow visual context to directly modulate structural choices, mirroring this human cognitive integration.

In contrast, the weaker correlations in dual-encoded models likely reflect architectural bottlenecks where visual and textual streams remain largely independent until late-stage fusion. While these models can still exhibit structural priming, the effect appears to be primarily driven by linguistic similarity rather than genuine visual-syntactic integration. This suggests that structural priming in dual models may be a byproduct of language model training rather than evidence of multimodal cognitive processing.

The unexpected finding that visual similarity correlates more strongly with structural preservation than semantic similarity challenges conventional assumptions about language processing hierarchies. We propose that visual scenes contain inherent compositional structures (spatial relationships, agent-patient configurations, event sequences) that naturally align with syntactic templates, providing more stable structural cues than abstract semantic content. This visual dominance may reflect the grounding problem in language models: visual features provide concrete anchors that help stabilize syntactic choices across different linguistic expressions of the same scene.

Our findings contribute to broader discussions about cognitively plausible AI architecture \citep{Lake_Ullman_Tenenbaum_Gershman_2017}. These results suggest that achieving human-like structural priming requires deep representational integration that allows cross-modal influences to emerge naturally—not merely processing multiple modalities independently. The failure of dual-encoded models to show strong visual-syntactic correlations, despite their multimodal capabilities, demonstrates that architectural choices fundamentally constrain whether human-like cognitive behaviors can emerge. This research bridges computational linguistics and cognitive science, offering computational validation of psycholinguistic theories while revealing how engineering decisions in AI systems can either facilitate or hinder natural cognitive processes.

\section{Conclusion}

We developed the first multimodal structural priming dataset named PRISMATIC, together with a SPI metric that evaluates models' priming effects without requiring standard answers. The SPI serves as a new standard for assessing how well machine predictions preserve syntactic structures from previous contexts. With controlling experiments, we constructed and tested MLLMs with two different encoding methods. Contrary to traditional beliefs, we found no significant statistical difference in syntactic priming ability between dual and fusion encoders, suggesting that different encoding methods have similar capabilities in preserving syntax.

However, correlation analysis revealed an interesting pattern: only fusion-encoded models showed a strong positive correlation between syntactic priming effects and image similarity, while in dual-encoding models, syntactic priming effects were unrelated to image similarity. This indicates that fusion encoding more closely resembles human psycholinguistic cognitive processes. However, the mechanism that enhances syntactic priming capabilities in dual encoding still requires further investigation; its priming performance is similar, but the underlying mechanism warrants further investigation.

\section*{Limitations}
Although we proposed a new dataset, a new metric, built models for controlled experiments, and observed a significant Pearson correlation coefficient, our research still has limitations. 

Due to the reliance on manual verification, our current dataset size is limited and insufficient for training. Despite Flickr30k's size of over 30,000 images and 150,000 captions, the dataset construction remains challenging due to the complexity of certain syntactic structures. In particular, finding images that can appropriately elicit specific complex types, such as mediopassive syntax,  limits our potential sample size.

While our SPI metric can effectively measure both positive and negative priming effects, it lacks the ability to assess situations where the model output completely deviates from the expected syntactic structure. Although we verify the image-text alignment via cosine similarity score, the SPI metric itself cannot directly assess the semantic consistency between the generated description and the visual content.

During the experiment, manual observation found that hallucinations occur in fusion encoding models more frequently, but the current evaluation method cannot quantify the level of the hallucinations, nor can we prove whether the hallucinations are caused by prime information.


\bibliography{acl_latex}

\appendix

\clearpage
\begin{appendix}


\section{Syntax Tree Representation}
\label{appendix:syntax_tree}
\begin{figure}[ht]
    \centering  
    \includegraphics[width=1.95\linewidth]{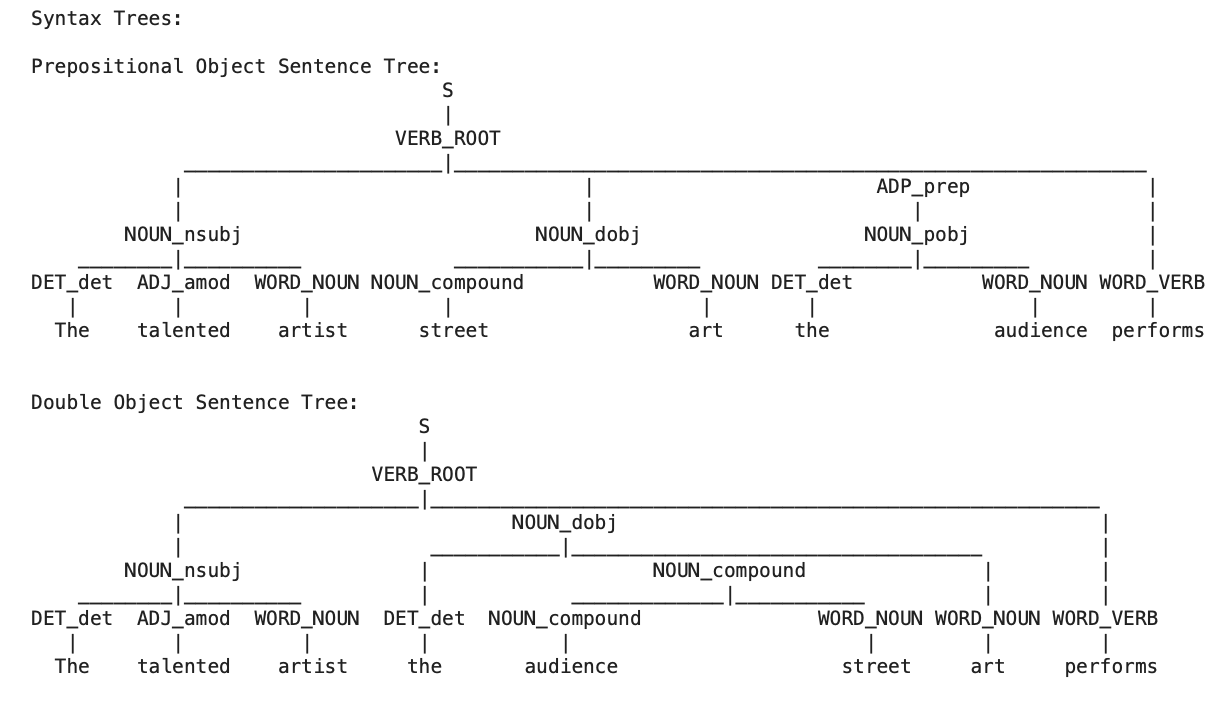}
    \caption{Syntax trees of a PO sentence and a DO sentence, illustrating structural differences.}
    \label{fig:parse_appendix}
\end{figure}

Syntax trees generated in Figure~\ref{fig:parse_appendix} show a comparison of the structure of positive prime and negative prime sentences.

\end{appendix}

\end{document}